\documentclass[10pt,twocolumn,letterpaper]{article}

\usepackage{cvpr}
\usepackage{times}
\usepackage{epsfig}
\usepackage{graphicx}
\usepackage{amsmath}
\usepackage{amssymb}
\usepackage{tabularx}
\usepackage{subfig}
\usepackage[all]{nowidow}
\usepackage[numbers,sort]{natbib}

\usepackage[pagebackref=true,breaklinks=true,letterpaper=true,colorlinks,bookmarks=false]{hyperref}

\cvprfinalcopy 

\def\cvprPaperID{4} 

\ifcvprfinal\fi
\begin{document}

\title{Event detection in coarsely annotated sports videos via parallel multi receptive field 1D convolutions}


\author{Kanav Vats \quad \text{Mehrnaz Fani} \quad \text{Pascale Walters} \quad \text{David A. Clausi} \quad  \text{John Zelek}\\
University of Waterloo \hspace{2cm}\\
Waterloo, Ontario, Canada\\
{\tt\small \{k2vats,mfani,pbwalter,dclausi,jzelek\}@uwaterloo.ca}}

\maketitle

\begin{abstract}

In problems such as sports video analytics, it is difficult to obtain accurate frame level annotations and exact event duration because of the lengthy videos and sheer volume of video data. This issue is even more pronounced in fast-paced sports such as ice hockey. Obtaining annotations on a coarse scale can be much more practical and time efficient. We propose the task of event detection in coarsely annotated videos. We introduce a multi-tower temporal convolutional network architecture for the proposed task. The network, with the help of multiple receptive fields, processes information at various temporal scales to account for the uncertainty with regard to the exact event location and duration. We demonstrate the effectiveness of the multi-receptive field architecture through appropriate ablation studies. The method is evaluated on two tasks - event detection in coarsely annotated hockey videos in the NHL dataset and event spotting in soccer on the SoccerNet dataset. The two datasets lack frame-level annotations and have very distinct event frequencies. Experimental results demonstrate the effectiveness of the network by obtaining a 55\% average F1 score on the NHL dataset and by achieving competitive performance compared to the state of the art on the SoccerNet dataset. We believe our approach will help develop more practical pipelines for event detection in sports video.

\end{abstract}

\section{Introduction}

\begin{figure*}[t]
\begin{center}
\includegraphics[width=\linewidth, height =.37\linewidth]{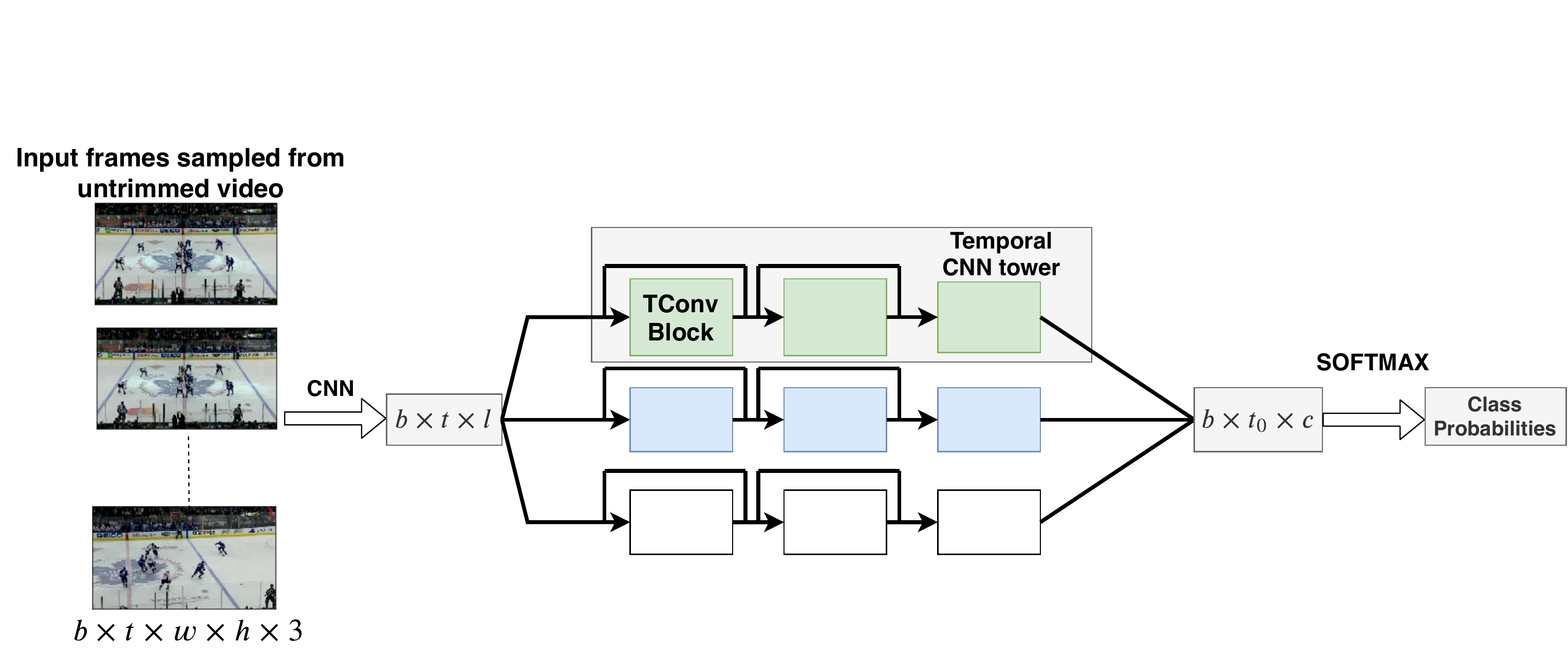}
\end{center}
   \caption{The overall network architecture. The network takes $t$ frames of dimension $w \times h \times 3$  sampled from untrimmed sports video as an input. A CNN extract $l$ dimensional features from the video frames. The $t \times l$ features are input into a three tower temporal convolutional network. Any two towers with nodes at a particular layer have different receptive fields. The output is $t_{0} \times c$ dimensional where $t_{0}$ is the number of contiguous events predicted for the $t$ frames and $c$ is the number of output classes. Here, $b$ denotes the batch size. Finally softmax function is applied to obtain probabilities.}
\label{fig:network}
\end{figure*}

Sports analytics has recently emerged as one of the major applications of computer vision. Various problems such as player tracking \cite{6619083}, sports broadcast video registration \cite{Chen_2019_CVPR_Workshops,8354144} and sports camera selection \cite{8658607} are being solved with the aid of computer vision. Event detection is a challenging problem when it comes to applications of computer vision in sports. This is because of the fast paced events in certain sports such as ice hockey and lack of publicly available datasets dedicated to sports.\par 
Most of the current papers in sports event detection take advantage of frame-level annotations. Despite the availability of a large number of sports videos on online platforms such as YouTube, frame-by-frame annotations are quite difficult to obtain. As such, it can be much easier to obtain coarser, second or minute-wise annotations. The downside of this is that the annotations will be coarse and approximate, which can cause problems in sports where events last for short time spans.  \par
In this paper, we introduce a practical paradigm for event detection in coarsely annotated untrimmed sports videos. To accomplish this, we introduce a multi-towered temporal 1D convolutional architecture for event detection. Video frames are input into a pretrained 2D CNN to obtain input feature vectors. The feature vectors are fed to the 1D convolutional towers. Each tower processes input feature vectors on different temporal scales with the help of varying temporal receptive fields. The activations from the parallel towers are finally added up to obtain class probabilities. The overall network architecture is shown in Fig. \ref{fig:network}.\par
We evaluate our methodology on the NHL dataset and SoccerNet dataset \cite{Giancola_2018_CVPR_Workshops}. The NHL dataset is a densely annotated dataset with a high event frequency where each second is annotated with an event. The frame level location and duration of the event is not defined. The SoccerNet dataset also presents a more practical scenario where soccer events are anchored to particular seconds (called spots) in the video. Hence, the two datasets represent two coarsely annotated datasets with exactly opposite event frequencies. For the high event frequency hockey dataset, the output node of each tower observes a different receptive field taking into account the uncertainty in the location and duration of the event in the coarsely annotated video. We experimentally demonstrate the effectiveness of our network architecture when compared to a fixed receptive field network with an appropriate ablation study. We obtain an F1 score of $\sim$ 55\% on the dataset.  We address the task of event spotting in the sparsely annotated SoccerNet dataset using our network and obtain competitive performance compared to the state of the art \cite{cioppa2019context}.


\section{Background}


\textbf{Video understanding}. Video understanding is one of the most important avenues for computer vision research. Action recognition \cite{Simonyan2014TwoStreamCN, lrcn2014, c3d, carreira2017i3d} and temporal event localization \cite{Chao2018RethinkingTF,Farha2019MSTCNMT,hussein2018timeception,Lei2018TemporalDR} are two major problems addressed in video understanding literature. Action recognition consists of recognizing actions from trimmed video clips. Various techniques such as two stream networks \cite{Simonyan2014TwoStreamCN}, 3D convolutions \cite{c3d} and recurrent neural networks \cite{lrcn2014} have been utilized for action recognition. Other popular works \cite{carreira2017i3d} in action recognition use two-stream inflated 3D convolutions obtained by pretraining 2D CNN filters on Imagenet \cite{imagenet_cvpr09}  and then inflating 2D filters to 3D by repeating weights depth wise. \par Temporal event localization \cite{Chao2018RethinkingTF,hussein2018timeception,Lei2018TemporalDR, Farha2019MSTCNMT} consists of locating the start and end frame for actions in untrimmed videos. Although \cite{Lei2018TemporalDR, Farha2019MSTCNMT} make use of temporal 1D convolutions, they however, require frame level annotations. Our work is related to weakly supervised approaches for temporal event localization \cite{Richard2017WeaklySA,autoloc}, since we estimate event locations in untrimmed videos without frame-level annotations. Our work is also in line with TAL-Net \cite{Chao2018RethinkingTF} and Timeception \cite{hussein2018timeception}. TAL-Net\cite{Chao2018RethinkingTF} ,based on the structure of Faster R-CNN, TAL-Net performs action recognition using multi-scale anchor proposals by suggesting segments from an untrimmed video with a small 1D CNN. Hussein \textit{et al.} \cite{hussein2018timeception} introduce Timeception layers for long-range complex action recognition. Timeception layers perform multi-scale temporal only convolutions with reduced complexity and can be used with either a 2D or 3D CNN backbone. The focus in this paper is on event localization in sports videos without frame level annotations, using ice hockey and soccer datasets having different event frequencies.  \\\\

 \textbf{Sports video analytics}.  Event recognition \cite{Tora2017, mehrasa2018deep, Piergiovanni_2019_CVPR_Workshops}, player level action recognition \cite{fani2017hockey, cai2019temporal, fani2019pose} and event detection \cite{Giancola_2018_CVPR_Workshops, mcnally2019golfdb, Victor_2017_CVPR_Workshops, cioppa2019context} in sport videos are some of the active research efforts computer vision. Tora \textit{et al.} \cite{Tora2017} predict hockey events using a single layered LSTM \cite{hochreiter1997long} architecture on top of a pre-trained AlexNet \cite{krizhevsky2012imagenet}. Mehrasa \textit{et al.} \cite{mehrasa2018deep} perform group activity recognition in hockey with the help of player trajectories using 1D convolutions. Fani \textit{et al.} \cite{fani2017hockey} recognize individual hockey player\rq{}s action type by estimating the pose of the player in each video frame, using a stacked hourglass network \cite{newell2016stacked}, without incorporating temporal information. Cai \textit{et al.} \cite{cai2019temporal} use the coordinates of the player's hockey stick as part of the pose of the  hockey players in each frame, in conjunction with optical flow in a two stream architecture. In another work, Fani \textit{et al.} \cite{fani2019pose} performs action recognition of individual soccer players from video by extracting the pose of the player, normalizing it and applying LSTM layers for capturing the temporal variation of the player\rq{}s pose during the action performance.

\begin{figure*}
	\begin{center}
		\subfloat[Towers $T_{1},T_{2}\,and\,T_{3}$]{
		    \includegraphics[width=0.33\linewidth]{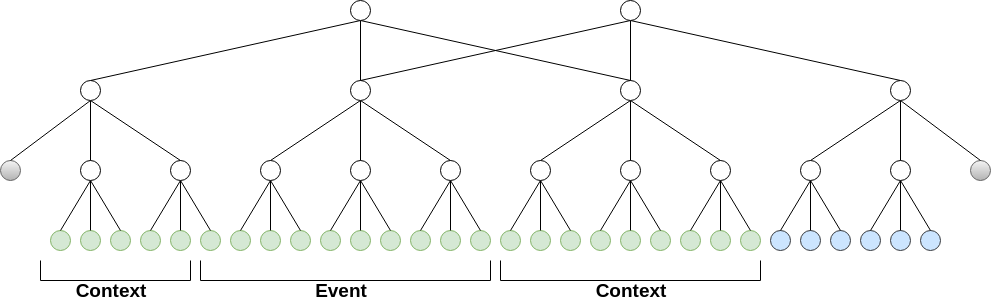}
		    \includegraphics[width=0.33\linewidth]{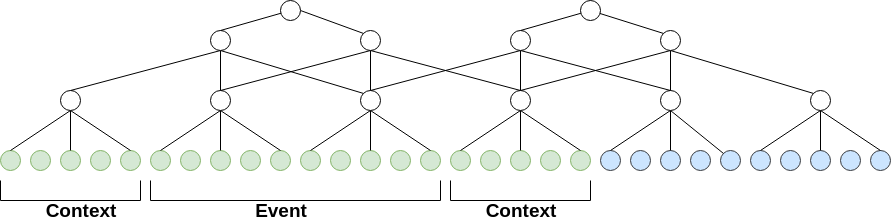}
		    \includegraphics[width=0.33\linewidth]{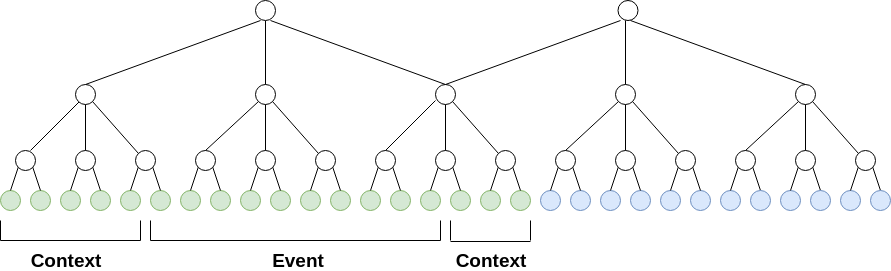}
		}
	
	\end{center}
	\caption{Illustration of temporal convolutional towers denoted by $T_{1}$, $T_{2}$ and $T_{3}$. The first block of each tower has a receptive field of 3, 5 and 2 respectively. The coloured nodes represent the input feature vectors. The two output nodes of each network corresponds to two contiguous events. The green colored nodes represent the input feature vectors in the receptive field of the first output node. The receptive field of a particular node covers the event itself and the context around it. Note that the receptive field of the output nodes in each tower is different. }
	\label{fig:detail_network}
\end{figure*}

Due to the lack of standard datasets for event detection in sports videos, many researchers generate their own datasets which usually have limited size, and are often not generalizable. To address this issue, Giancola \textit{et al.} \cite{Giancola_2018_CVPR_Workshops} introduce SoccerNet, a benchmark for event spotting in soccer videos. This benchmark, which is generated for the purpose of localizing very sparse events within long videos, spots three main event types in 500 soccer games. McNally \textit{et al.} \cite{mcnally2019golfdb} introduce a benchmark database for detecting eight events in the golf swing, named GolfDB consisting of 1400 golf swing videos. \par
The above research, with the exception of Giancola \textit{et al.} \cite{Giancola_2018_CVPR_Workshops}, either utilize frame level annotations \cite{Tora2017, mehrasa2018deep, cai2019temporal, fani2019pose, fani2017hockey,mcnally2019golfdb, Victor_2017_CVPR_Workshops} or classify trimmed video clips \cite{fani2017hockey, fani2019pose, cai2019temporal, Piergiovanni_2019_CVPR_Workshops,k2crv}. Here, we focus more on the practical case where frame-by-frame video annotations are not available in untrimmed video datasets of different event frequencies.

\section{Methodology}
The proposed approach for event detection is explained in Subsection 3.1. The designed network  is shown in Figure \ref{fig:network} and explained in Subsection 3.2.

\subsection{Proposed Approach}
Events in sport videos occur at varying temporal scales. For instance, in hockey, events such as `shots', usually occur in a shorter time span than an event like a `faceoff'. To take this factor into account, we employ 1D CNNs of varying kernel sizes and receptive fields. The information from these parallel 1D CNNs is fused to obtain event class probabilities. The network architecture is described in the next section.  

\subsection{Network Architecture}
To detect events in untrimmed sport videos, we make use of a multi-tower architecture. The towers represent temporal 1D CNNs with different receptive fields. The output node gives the probability of an event occurring in the video. The network architecture is illustrated in Figure \ref{fig:network}.

The input to the network is a sequence of $t$ frames $ \{I_{k} \in R^{w \times h \times 3} : k \in \{ 1,2,...,t \} \} $ sampled uniformly at a frame rate of $f$ frames per second from an untrimmed sports video. The images are passed through a 2D CNN in order to obtain features $F_{k} \in R^{l} $ from an intermediate layer. Separate 1D convolution towers of varying kernel sizes (or varying receptive fields) are applied on top of the features $F_{k}$. The kernel size and stride of CNN filters in the towers is chosen such that for each tower, a node in a particular layer has a different receptive field than the corresponding node in other towers. We incorporate contextual features such that the network sees what happens immediately before and after an event.

\begin{figure}[t]
\begin{center}
\includegraphics[width=.4\linewidth, height=6cm]{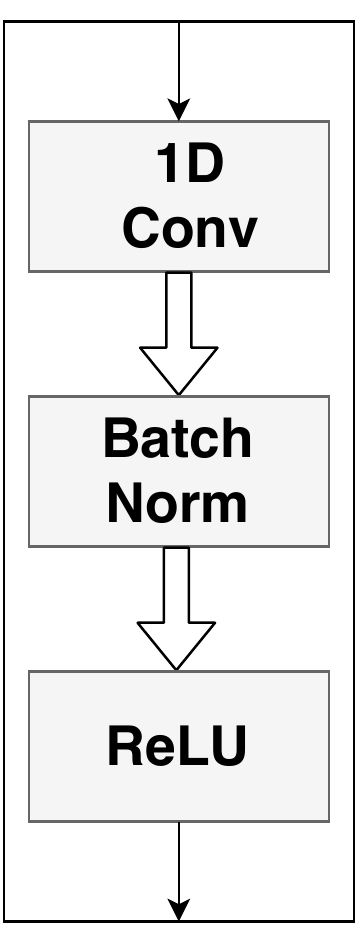}
\end{center}
  \caption{TConv block consists of a 1D Convolutional layer, Batch Normalization layer and ReLU activation function. Since $t^{'} <t$, the residuals are sliced symmetrically for making the skip connection. $b$ and $l$ denote the batch size and feature dimension respectively.}
\label{fig:TCN_block}
\end{figure}

Each 1D convolutional network is composed of a number of 1D convolution blocks named TConv block, of structure illustrated in Figure \ref{fig:TCN_block}. The blocks are composed of a 1D convolutional layer followed by a batch normalization layer ~\cite{Ioffe:2015:BNA:3045118.3045167} and ReLU non-linearity. In addition, skip connections ~\cite{HeResnet} are also added by slicing the input to match the output. While adding skip connections, downsampling is done in case dimensionality of input is different than that of the output. The output  $ O_{t} \in R^{t_{0} \times c}$  where $t_{0}$ is the number of contiguous events predicted and $c$ represents the number of output classes. The output of 1D convolutional towers are added together and  then softmaxed to obtain event class probabilities.

\section{Datasets Used} \label{sec:datasets}

We have used two different datasets for our experiments. The two datasets have a large variation in event frequency. 

\subsection{NHL dataset}
\begin{figure*}
	\begin{center}
		\subfloat[Faceoff]{
		    \includegraphics[width=0.23\linewidth]{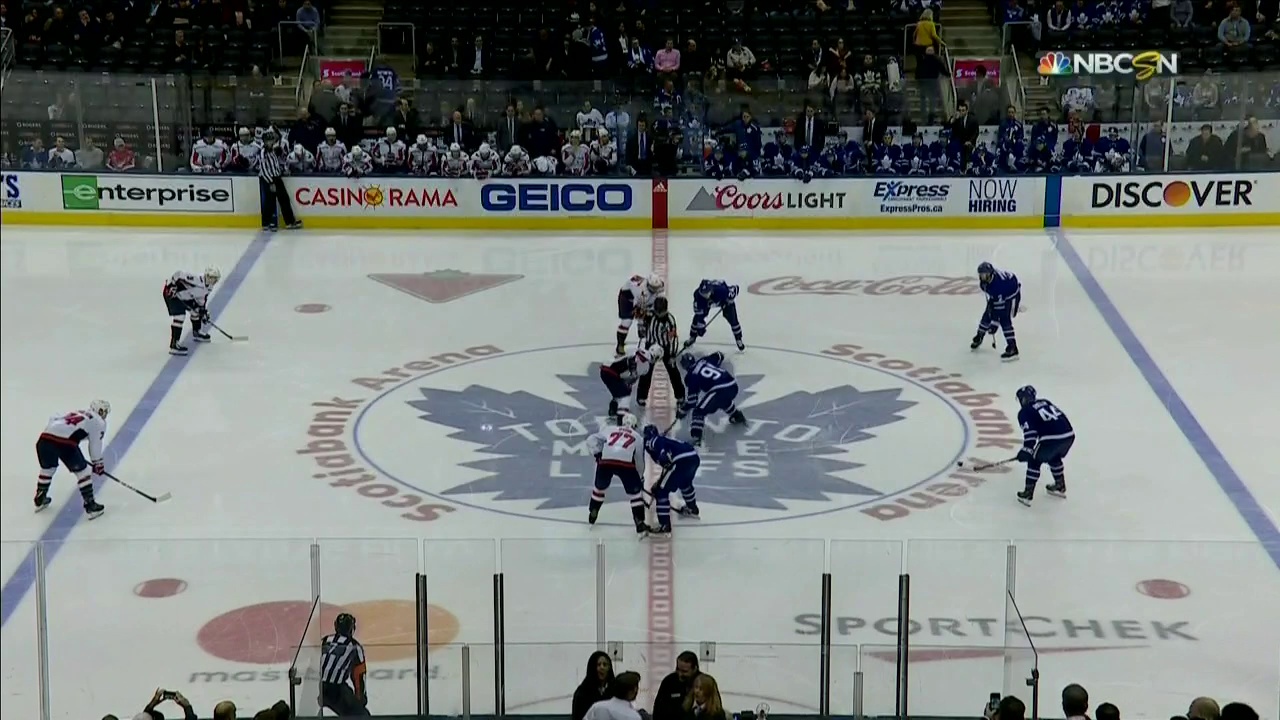}
		}
		\subfloat[Shot]{
		    \includegraphics[width=0.23\linewidth]{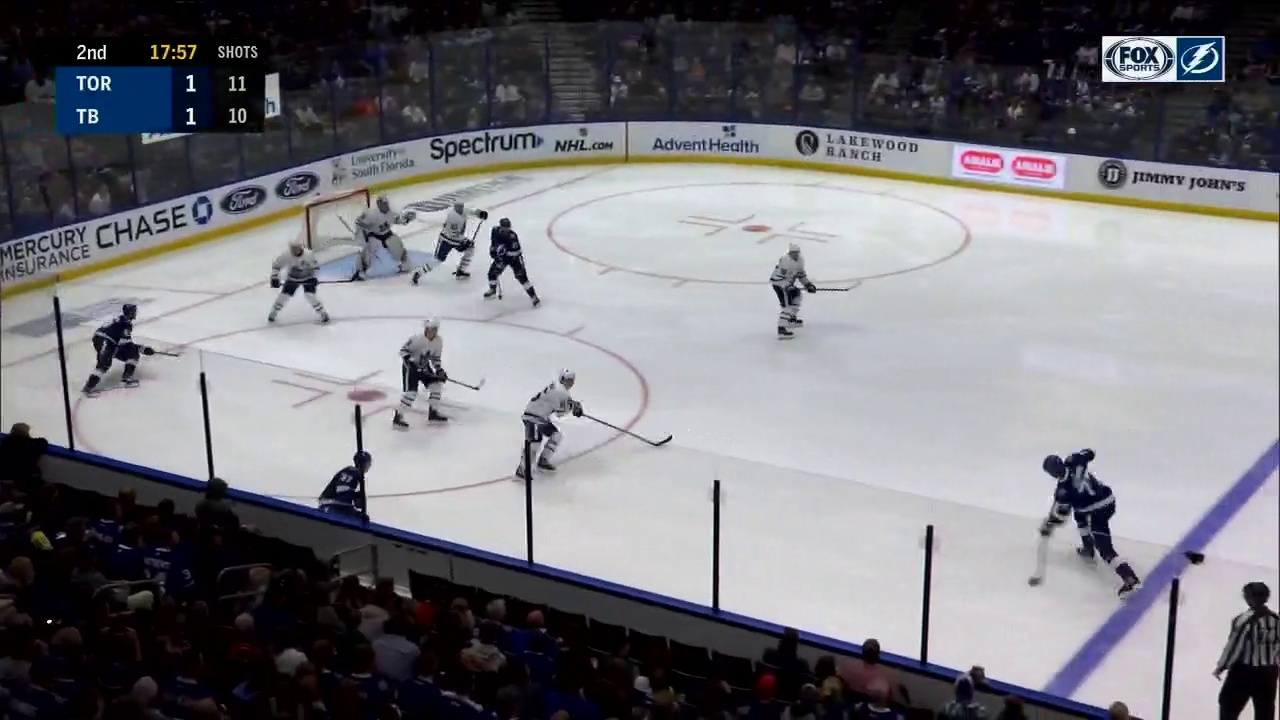}
		}
		\subfloat[Advance]{
		    \includegraphics[width=0.23\linewidth]{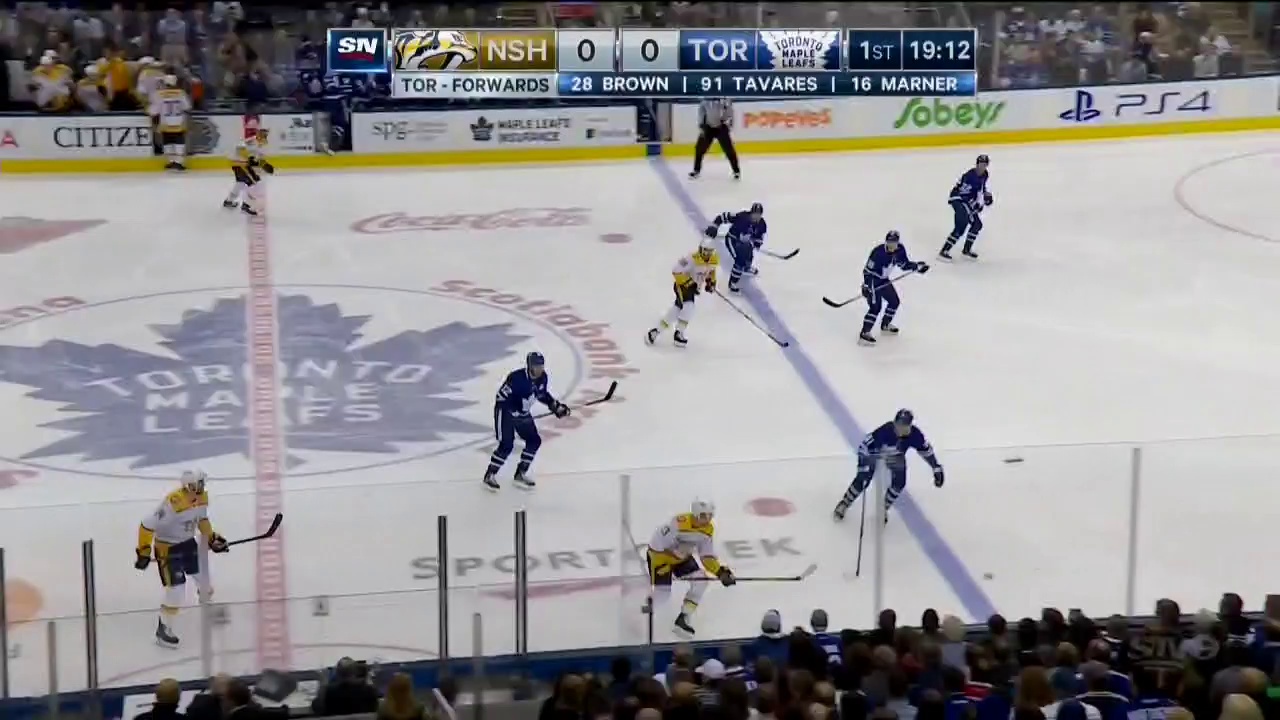}
		}
		\subfloat[Play]{
		    \includegraphics[width=0.23\linewidth]{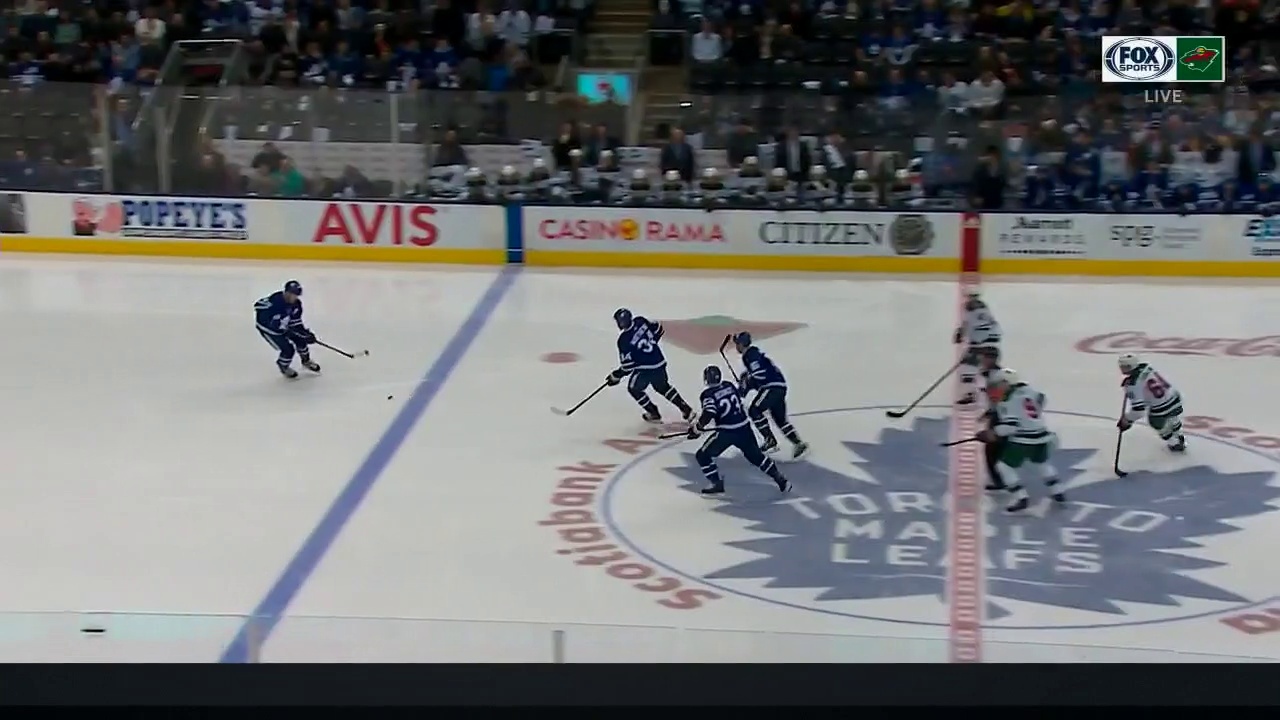}
		}
	\end{center}
	\caption{Examples of frames from each of the annotated events}
	\label{fig:example_events}
\end{figure*}

The NHL dataset consists of 10 NHL games of three periods each, with separate 60 fps videos for all game periods. The videos come from broadcast footage and include advertisements, replays, shots of varying range, and overlayed graphics. Each video has a spatial resolution of $1280 \times 720$ pixels. The videos are annotated with one second resolution, whereby the event is expected to occur within the one second interval. The events are annotated with at least one of the following labels: \textbf{Faceoff}, \textbf{Shot}, \textbf{Advance}, or \textbf{Play}. Table \ref{table:event_desc} gives descriptions of the event types and Figure \ref{fig:example_events} shows example frames. The dataset is heavily imbalanced with Play consisting of $ \sim 80$\% of all events. In some cases, an annotated Play event may overlap with another event of another type. In this case, the time frame is simply assigned to the non-Play event without affecting the data distribution. The dataset contains 589 Faceoffs, 1,062 Shots , 1,306 Advance and 11,116 Play events. The dataset has a high event rate of one event every 4.5 seconds. The actual event rate is higher when excluding advertisements.

\begin{table*}
    \centering
    \caption{Event descriptions in the NHL dataset.}
    \setlength{\tabcolsep}{0.2cm}
    \begin{tabularx}{\textwidth}{|l|X|}
        \hline
        \textbf{Event} & \textbf{Description} \\
        \hline
        Faceoff & The puck is dropped between the sticks of two opposing players \\
        Shot & A player attempts to shoot the puck on goal \\
        Advance & A player moves the puck into or out of the defensive or offensive zone without an intended recipient (e.g., dump in, clearing attempt) \\
        Play & A player moves the puck with an intended recipient (e.g., pass, stickhandle) \\
        \hline
    \end{tabularx}
    \label{table:event_desc}
\end{table*}

The annotations are approximate and coarse. The frames that represent an event usually span a fraction of a second and may actually be present outside the annotated one second window. This also means that the exact frame-wise location of the event is not defined. The annotations were collected manually and this annotation scheme is more practical than frame-level annotations, which are very difficult and time consuming to obtain. The dataset is split such that nine games are used for training, and one period and two periods from the remaining game are used for validation and testing, respectively.

\subsection{SoccerNet Dataset}
The SoccerNet dataset \cite{Giancola_2018_CVPR_Workshops} is composed of 500 soccer games from the main European Championships from three seasons with a total duration of 764 hours. The events are categorized into three categories: \textbf{Yellow/Red Card}, \textbf{Goal}, or \textbf{Substitution}. The dataset is very sparse such that it contains an average of one event every 6.9 minutes, making the task of event localization difficult. For each event, temporal anchors of one second resolution are obtained according to well-defined soccer rules. The 500 games are randomly split into 300 games for training, 100 for validation and 100 for testing. We use the same split as Giancola \textit{et al.} \cite{Giancola_2018_CVPR_Workshops} for our experiments. PCA reduced 512 dimensional backbone features are provided corresponding to ResNet \cite{HeResnet}, C3D \cite{c3d} and I3D \cite{carreira2017i3d} networks. The features were extracted every 0.5 seconds from the video.

\begin{figure*}
	\begin{center}
		\subfloat[Ground Truth: Play, Predicted: Advance]{
		    \includegraphics[width=\linewidth]{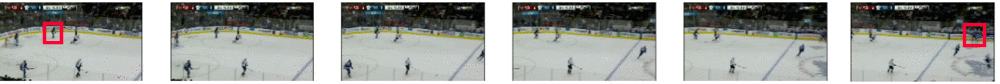}
		}\\
		\subfloat[Ground Truth: No event, Predicted: Faceoff]{
		\includegraphics[width=\linewidth]{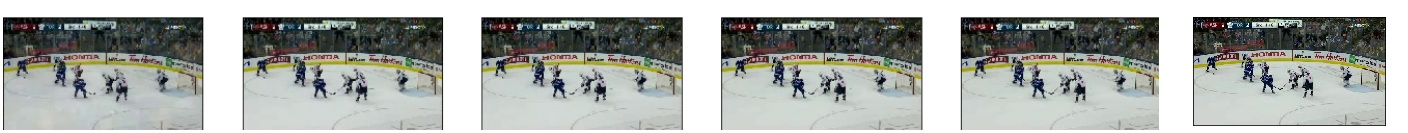}
		}
	\end{center}
	\caption{Two of the common kinds of failure cases. In (a), the network predicts a long pass as an Advance event. The red boxes denote the two players between whom the pass is being made. In (b),the faceoff is called off before actually happening, usually due to a faceoff violation on the play.  }
	\label{fig:failure_cases}
\end{figure*}

\section{Experiments}
We perform experiments on the NHL and the SoccerNet datasets mentioned above. We introduce the task of event detection in coarsely annotated videos using the NHL dataset and address the task of event spotting on the SoccerNet dataset \cite{Giancola_2018_CVPR_Workshops}.

\subsection{Event detection in coarsely annotated NHL videos}

\subsubsection{Objective}
The objective of this task is to detect events from coarsely annotated untrimmed hockey videos.

\subsubsection{Experiment Settings}

The videos are first downsampled to a resolution of $284 \times 160$ ($w = 284, h = 160$) pixels such that the initial aspect ratio is maintained. Images are sampled uniformly at a rate of 10 frames per second. We sample a total of $t=30$ frames for a period of 3 seconds. The network architecture is designed such that the number of output nodes are two i.e., we predict the output for $t_{0} =2$ contiguous seconds (detail in Figure \ref{fig:detail_network}). This is because, the NHL dataset is quite dense and two different events for instance, shot and play can occur consecutively. We subtract ImageNet \cite{imagenet_cvpr09} mean and divide by ImageNet standard deviation for normalization. A MobileNetV2 \cite{mobilenet} pretrained on ImageNet is used to extract features from the video. Global average pooling is performed on the final layer of MobileNetV2 to obtain $l =1280$ dimensional features. The network architecture used to process the 1280 dimensional features is shown in Table \ref{table:hockey-net}. We use a three towered architecture with the first block of the towers having an effective receptive field of 2,3 and 5 respectively. Random horizontal flipping is used for data augmentation.

Since the background covers a major proportion of the video, in order to handle the heavy class imbalance in the dataset, we explicitly control the event sampling such that the background events are sampled with a probability of $p_0$ and the events are sampled with probability $1 - p_{0}$. The value of $p_{0}$ is empirically chosen as 0.2. This is done to ensure that the training batches contain an even distribution of all $c =5$ event classes (including background), without which the model finds it difficult to converge. Further, in all experiments, a weighted cross entropy loss is used with play event and background assigned a weight of .05 and .033 respectively and the rest of the classes are assigned a weight of 1 each.  Adam optimizer is used with an initial leaning rate of 0.001. The training is done on an Nvidia GTX 1080 Ti GPU.


\begin{figure}[t]
\begin{center}
\includegraphics[width=6cm, height=5cm]{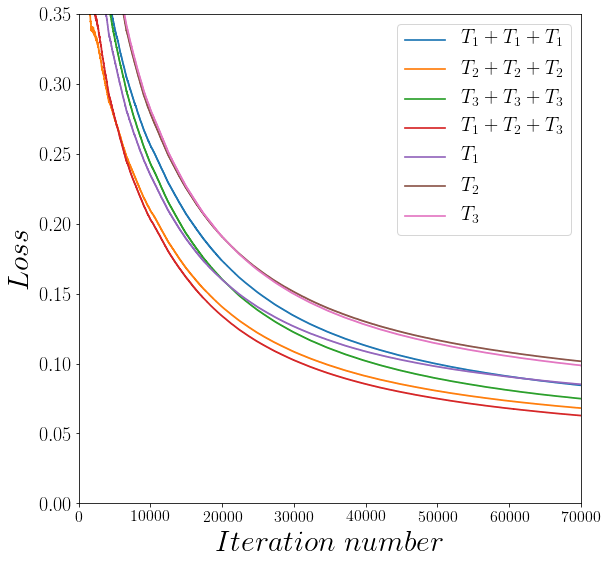}
\end{center}
   \caption{Training loss vs Number of Iterations for multi tower settings. The configuration $T_{1}+T_{2}+T_{3}$ (red curve) attains the lowest loss values as compared to the other configurations. }
\label{fig:losses}
\end{figure}

\subsubsection{Post processing}

During evaluation/testing phase, we apply the network in a sliding window fashion with a stride of one second on the untrimmed video. Since the testing is done with a stride of one second, each event is predicted twice. We take the maximum confidence of these two predictions. Furthermore, we take advantage of the fact that events such as Faceoff, Advance and Shot are extremely less likely to occur consecutively. If the network predicts one of these event $n$ times in a row where $n > 1$, we only consider the prediction with maximum confidence. The rest of the predictions are assigned the prediction with second largest confidence. This leads to an average improvement of 2-3 \% in F1 score values.

\subsubsection{Results and Analysis}

A predicted event for a one second interval is considered correct if it is within one second of any ground truth. It can be understood as accuracy within a tolerance of $\delta = 1$ second. We calculate the precision, recall and F1 score for each class according to the above definition. Table \ref{table:pr_values} shows the F1 scores for each class.

Faceoff and Play events have the highest F1 score whereas advance event has the lowest F1 score values (65.62 and 65.04 respectively). From the low precision value (32.23) of Advance events, it can be concluded that other events are often mistaken for Advance events. A common observation is that long passes (Play events) are often mistaken as Advance events. Another issue arises when a faceoff is called off before actually happening, usually due to a faceoff violation on the play. The model, in this scenario, gets enough spatiotemporal information to classify it as a Faceoff event because of which false alarms are generated. These kind of failure cases are illustrated in Fig. \ref{fig:failure_cases}. Also, many times, the hockey players are occluded by the near boards of the rink.

\begin{table}[!t]
    \centering
    \caption{Precision, Recall and F1 score values for the network for the NHL dataset}
    \footnotesize
    \setlength{\tabcolsep}{0.2cm}
    \begin{tabular}{c|c|c|c|c|c}\hline
       & Faceoff & Shot & Advance  & Play & Average \\\hline\hline
       Precision & 77.78 & 52.74 & 32.23 & 51.42 & 53.54\\ 
       Recall & 56.76 & 44.86 & 44.88 & 88.48 & 58.74 \\ 
       F1 score & 65.62 & 48.49 & 40.70 & 65.04 & 54.97\\ 
    \end{tabular}
    \label{table:pr_values}
\end{table}

\subsubsection{Ablation studies}

Table \ref{table:ablation_table} shows the performance of the individual towers in the first three rows. For comparing inherent network performance, comparison values are based on network outputs excluding post processing. Repeating the same tower three times and jointly training the three towers performs at-least as good or better than a single tower as demonstrated by the higher F1 score of $T_{2}+T_{2}+T_{2}$ and $T_{3}+T_{3}+T_{3}$ configuration than their single tower counterparts. This is due to the increase in representational power from the increase in network capacity. This is further seen in the training loss of the respective models (Fig. \ref{fig:losses}).\par
We perform ablation experiments on the number of temporal convolutional towers. The purpose of the study is to demonstrate that the increase in performance is not merely because of the increase in network capacity. Table \ref{table:ablation_table} demonstrates that using towers with different receptive fields is important. Repeating the same tower three times, although increases parameters, but does not improve performance when compared to a multi receptive field network. This is evident by the highest F1 scores obtained by the three different receptive field setting (51.56). This is also demonstrated by the lowest training loss value of $T_{1}+T_{2}+T_{3}$ setting in Fig \ref{fig:losses}. Repeating a fixed receptive field tower three times has one of the two following effects: either redundant information is provided to the network, (which means the receptive field is too large for an event), or less information is provided if the receptive field is too small, (which results in lower accuracy).
\begin{figure*}[t]
\begin{center}
\includegraphics[width=.8\linewidth, height =.30
\linewidth]{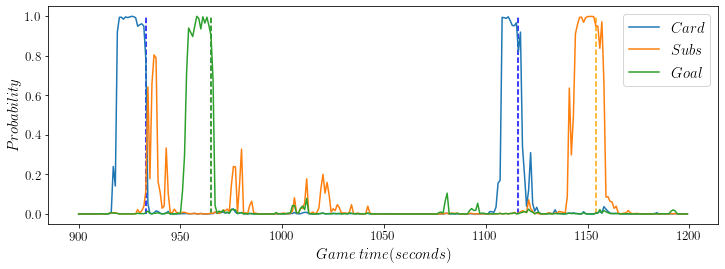}
\end{center}
  \caption{Event detection probability vs game time plot for a 5 minute interval in the second half of 2016-2017 season UEFA Champions league Barcelona vs PSG game(6-1). The vertical dashed lines denote the ground truth spot timings. The network generates clean proposal segments for each event type. The high substitution probabilities between 900 - 1050 second occur during replay-highlights of card and goal. The replay-highlights are a card and substitution are often similar, where camera is focused on a single player leading to false positives and lower precision for these events. }
\label{fig:probs_game_time}
\end{figure*}

\begin{table}[!t]
    \centering
    \caption{Comparison of various 3-tower configurations for the NHL dataset. $T_{1}$, $T_{2}$ and $T_{3}$ represent temporal convolutional towers having different receptive fields. Highest F1 score values are obtained on using towers with different receptive fields. }
    \footnotesize
    \setlength{\tabcolsep}{0.15cm}
    \begin{tabular}{c|c|c|c|c|c}\hline
  
      Network type & Faceoff & Shot & Advance  & Play & Avg F1  \\\hline\hline
       $T_{1}$ & 54.32 & 44.44 & 35.13 & 64.36 & 49.56\\ 
        $T_{2}$ & 45.98 & 41.32 & 23.91 & 63.79 & 43.75\\
        $T_{3}$ & 50.70 & 40.81 & 30.34 & 64.55 & 46.60 \\
     
       $T_{1} + T_{1} + T_{1}$ & 49.11 & 44.77 & 34.41 & \textbf{65.72} & 48.50 \\
       $T_{2} + T_{2} + T_{2}$ & 55.42 & \textbf{48.02} & 33.12 & 65.69 & 50.56 \\
       $T_{3} + T_{3} + T_{3}$ & 49.35 & 43.83 & 30.63 & 64.48 & 47.07\\
         $T_{1} + T_{2} + T_{3}$ & \textbf{56.76} & 45.59 & \textbf{38.86} & 65.18 & \textbf{51.60} 
       
    \end{tabular}
    \label{table:ablation_table}
\end{table}

\begin{table}[!t]
    \centering 
    \caption{Network architecture for the NHL dataset. Each column denotes a temporal convolution tower $T_{k}$. k,s,d and p denote kernel size, stride, dilation and padding respectively}
    \footnotesize
    \setlength{\tabcolsep}{0.2cm}
    \begin{tabular}{c|c|c}
    \hline $T_{1}$ & $T_{2}$ & $T_{3}$ \\\hline\hline
     k=3, s=3,d=1,p=0 & k=3, s=5,d=2,p=0  & k=2, s=2,d=1,p=0 \\
     Batch Norm 1D  & Batch Norm 1D & Batch Norm 1D \\
     ReLU & ReLU & ReLU \\ \hline
     k=3, s=3, d=1,p=1 & k=3, s=1,d=1,p=0  & k=3, s=3,d=1,p=0 \\
      Batch Norm 1D   & Batch Norm 1D  & Batch Norm 1D  \\
     ReLU & ReLU & ReLU \\ \hline
     k=3, s=1, d=1, p=0 & k=2, s=2,d=1,p=0  & k=3, s=2,d=1,p=0  \\
      Batch Norm 1D   & Batch Norm 1D  & Batch Norm 1D  \\
     ReLU & ReLU & ReLU \\ \hline
     \multicolumn{3}{c}{Sum} \\ \hline
     \multicolumn{3}{c}{Softmax} \\ \hline
    \end{tabular}
    \label{table:hockey-net}
\end{table}

\subsection{Event spotting in soccer}

\subsubsection{Objective}
The objective of this task is to find the anchors of soccer events in soccer game videos. We demonstrate the effectiveness of our approach by achieving competitive performance compared to the state of the art.
\subsubsection{Experiment Settings}

Instead of a two step approach used by Giancola \textit{et al.} \cite{Giancola_2018_CVPR_Workshops}, consisting of classification and then spotting, we train our model directly on the spotting task.  The model is trained on 15 second windows consisting of $t = 30$ features (features are extracted at 2fps from the video).
We again use the three tower architecture used in the NHL dataset. However, the model now outputs a single node ($t_{0} = 1$) representing the probability of the event. This is done by simply averaging the output of of final two nodes of the model used in the NHL dataset. This is done because, unlike the NHL dataset, the SoccerNet dataset is quite sparse and it can be safely assumed that a single event occurs in the 15 second interval.  The ground truth anchor is kept at the center of the sampled window of 15 second. 

On the testing data, we slide the network with a stride of one second in order to obtain event probabilities for one second resolution. As per Giancola \textit{et al.} \cite{Giancola_2018_CVPR_Workshops} we use a watershed method to generate segment proposals and use the center time in the segment to define the spotting candidate. \par
To handle the dataset imbalance resulting from the addition of the background class, we control event sampling with the value of parameter $p_{0} = 0.6$. A weighted cross entropy loss is used where the background class is given a weight of 0.33 and rest of the classes given a weight of 1 each.
The training is done with a batch size of $b = 120$. For data augmentation, we tried shifting the windows containing events by a random offset $s \in [-7.5,7.5]$ seconds from the event anchor, which, however did not bring any accuracy improvement. Adam optimizer is used with an initial leaning rate of 0.001. The training is done on an Nvidia GTX 1080 Ti GPU.

\begin{figure}[t]
\begin{center}
\includegraphics[width=4cm, height=3.5cm]{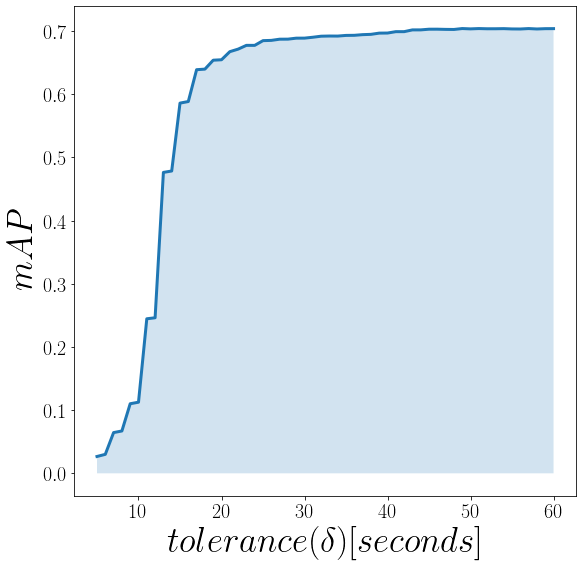}
\end{center}
   \caption{mAP as a function of tolerance $\delta$. The model obtains an average mAP of 60.1\%. }
\label{fig:auc}
\end{figure}

\subsubsection{Results and Analysis}

Giancola \textit{et al.} \cite{Giancola_2018_CVPR_Workshops} define the task of event spotting as finding the anchor time, called spot candidate that identifies the location of an event. A candidate spot is defined as positive if it lands within a tolerance $\delta$ around the ground truth anchor. Intuitively, the closer the candidate to a target, the better is the spotting performance. Mean average precision (mAP) is calculated for a given tolerance $\delta$. The accuracy metric is the average mAP between $\delta = 5$ to $\delta = 60$ seconds.
 \par
 
 \begin{figure}[t]
\begin{center}
\includegraphics[width=4cm, height=3.5cm]{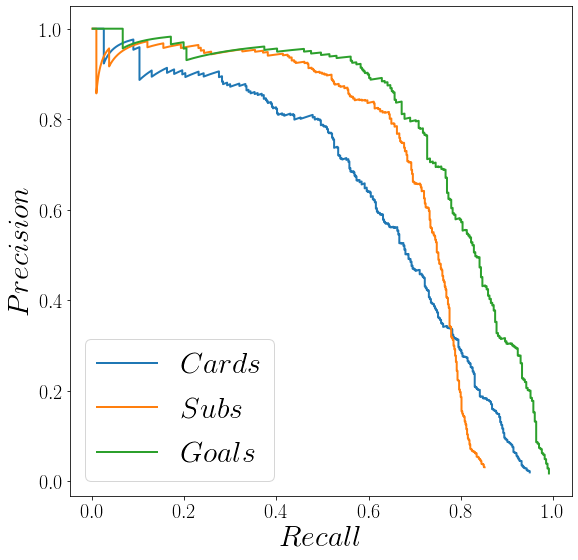}
\end{center}
   \caption{ Precision recall curves corresponding to the best performing model. }
\label{fig:pr_curve}
\end{figure}

\begin{table}[!t]

    \centering
    \caption{Class wise mAP scores for the SoccerNet dataset. Our method outperforms \cite{Giancola_2018_CVPR_Workshops} on all the classes}
    \footnotesize
    \setlength{\tabcolsep}{0.2cm}
    \begin{tabular}{c|c|c|c|c}\hline
  
       & Cards & Subs & Goals & Average \\\hline
       Giancola \textit{et. al} \cite{Giancola_2018_CVPR_Workshops}  & 52.1 &   59.3 & 73.0 & 61.5\\
       \textbf{Ours} & 63.1 & 69.1 &  79.0 & 70.4  \\ 
    \end{tabular}
    \label{table:SoccerNet_per_class}
\end{table}

  Giancola \textit{et. al.} \cite{Giancola_2018_CVPR_Workshops} showed that the I3D and C3D features already include temporal information, further incorporating these features in a temporal architecture leaves them redundant. Therefore, we use ResNet features in our experiments. Table \ref{table:SoccerNet_per_class} shows the highest per-class  mAP of the network (corresponding to $\delta = 60$ seconds) compared with  Giancola \textit{et. al}. Goal events are the easiest to spot obtaining an mAP of 79\%. Card events are the most difficult to spot with an mAP of 63.1\%. Figure \ref{fig:pr_curve} shows the corresponding precision-recall curves for the three classes. \par
 Figure \ref{fig:probs_game_time} shows the event probability vs game time plot for one of the soccer games in the test set.  The network generates clean proposal segments for each event type. A reason why the mAP for substitution and cards is low is because the replay-highlights of a card and substitution are often similar, where camera is focused on a single player leading to false positives and lower precision for these events. An example of this can be seen in Figure \ref{fig:probs_game_time} with significant value for substitution probability after the first card event.     \par
Figure \ref{fig:auc} shows the mAP vs tolerance($\delta$)  curve for tolerance between $\delta = 5$ to $\delta = 60$. From the shape of the curve, the mAP decreases almost linearly for tolerance below which the model was trained on i.e. 15 seconds. Around 60 second tolerance, the mAP saturates to $\sim$ 70 \%. We obtain an average-mAP of 60.1 \% averaged over the tolerances which exceeds  Giancola \textit{et. al} (49.7\%) by 10.4\% (Table \ref{table:SoccerNet_feature_type}). We argue that this is because our approach is able to understand the temporal aspect of the game better when compared to the two step NetVLAD \cite{Arandjelovic16} pooling (64 clusters) and Resnet152 \cite{HeResnet} based classification-detection approach used in Giancola \textit{et. al}. \cite{Giancola_2018_CVPR_Workshops}. Cioppa \textit{et. al.} \cite{cioppa2019context} recently introduced a context aware loss function for soccer action spotting using a combination of segmentation loss followed by an iterative matching procedure and a separate spotting loss. Our work achieves a competitive performance, with a difference of 2.4\% mAP (Table \ref{table:SoccerNet_feature_type}) and outperforms the ablation study baselines in Cioppa \textit{et. al.} \cite{cioppa2019context}, by using a much simpler network/approach using cross entropy loss function. 

\begin{table}[!t]

    \centering
    \caption{mAP scores for the soccer action spotting task. Our work achieves competitive results compared to the state of the art. }
    \footnotesize
    \setlength{\tabcolsep}{0.2cm}
    \begin{tabular}{c|c}\hline
  
       Method & mAP  \\\hline
           Giancola \textit{et. al} (5s) \cite{Giancola_2018_CVPR_Workshops} &  34.5\\
       Giancola \textit{et. al} (20s) \cite{Giancola_2018_CVPR_Workshops}  &  49.7\\
      Giancola \textit{et. al} (60s) \cite{Giancola_2018_CVPR_Workshops}  &  40.6\\
      Cioppa \textit{et. al} \cite{cioppa2019context} & 62.5 \\
       \textbf{Ours} &  60.1  \\ 
    \end{tabular}
    \label{table:SoccerNet_feature_type}
\end{table}

\section{Conclusion and Future Work}

In this paper, we address the difficulty of obtaining frame-level annotations in sport event detection. We introduce a multi-scale temporal 1D convolutional network for detecting events in two coarsely annotated datasets of completely different event frequencies. The results obtained on the SoccerNet dataset are more impressive than the hockey results. A reason for this is that the events in hockey are much more fast paced and frequent as compared to soccer, making the hockey dataset more challenging. Future work will be focused on taking advantage of player level contextual features, hockey puck position and game audio for the task of event detection.
\section{Acknowledgment}
This work was supported by Stathletes through the Mitacs Accelerate Program and Natural Sciences
and Engineering Research Council of Canada (NSERC).

{\small
\bibliographystyle{ieee_fullname}
\bibliography{cvsports}
}

\end{document}